\documentclass[conference]{IEEEtran}
\IEEEoverridecommandlockouts
\usepackage{cite}
\usepackage{amsmath,amssymb,amsfonts}
\usepackage{algorithmic}
\usepackage{graphicx}
\usepackage{textcomp}
\usepackage{xcolor}
\usepackage{lipsum}  
\usepackage{nidanfloat}
\usepackage{mathtools}
\usepackage{caption} 
\usepackage{float}
\usepackage{placeins}

\usepackage{subcaption}
 \usepackage[compact]{titlesec}
    \titlespacing{\section}{0pt}{2ex}{1ex}
    \titlespacing{\subsection}{0pt}{1ex}{0ex}
    \titlespacing{\subsubsection}{0pt}{0.5ex}{0ex}
{\subsection*{\normalsize\sagesf\bfseries Acknowledgements}\begin{refsize}\noindent #1}%
{\end{refsize}}
{\subsection*{\normalsize\sagesf\bfseries Funding}\begin{refsize}\noindent #1}%
{\end{refsize}}

\def\BibTeX{{\rm B\kern-.05em{\sc i\kern-.025em b}\kern-.08em
    T\kern-.1667em\lower.7ex\hbox{E}\kern-.125emX}}
\begin{document}
\title{Wi-Closure: Reliable and Efficient Search of
Inter-robot Loop Closures Using Wireless Sensing %in aliased environments
\author{Weiying Wang$^{1}$, Anne Kemmeren$^{2}$, Daniel Son$^{1}$, Javier Alonso-Mora$^{2}$, Stephanie Gil$^{1}$}

\thanks{$^{1}$ John A. Paulson School of Engineering and Applied Sciences, Harvard University, Allston, MA 02134, USA
        }%{\tt\small b.d.researcher@ieee.org}}% 150 Western Ave, Allston, MA 02134

\thanks{$^{2}$ Faculty of Mechanical, Maritime and Materials Engineering,
        Technical University of Delft, 2628 CD Delft, The Netherlands}
        %{\tt\small albert.author@papercept.net}}% Mekelweg 2 2628 CD Delft
}

\maketitle
\begin{abstract} In this paper we propose a novel algorithm, \textit{Wi-Closure}, to improve computational efficiency and robustness of loop closure detection in multi-robot SLAM. Our approach decreases the computational overhead of classical approaches by pruning the search space of potential loop closures, prior to evaluation by a typical multi-robot SLAM pipeline. \textit{Wi-Closure} achieves this by identifying candidates that are spatially close to each other by using sensing over the wireless communication signal between robots, even when they are operating in non-line-of-sight or in remote areas of the environment from one another. We demonstrate the validity of our approach in simulation and hardware experiments. Our results show that using \textit{Wi-closure} greatly reduces computation time, by $54\%$ in simulation and by $77\%$ in hardware compared, with a multi-robot SLAM baseline. Importantly, this is achieved without sacrificing accuracy. Using \textit{Wi-closure} reduces absolute trajectory estimation error by $99\%$ in simulation and $89.2\%$ in hardware experiments. This improvement is due in part to \textit{Wi-Closure}'s ability to avoid catastrophic optimization failure that typically occurs with classical approaches in challenging repetitive environments.

\end{abstract}

% \begin{IEEEkeywords}
% keywords\end{IEEEkeywords}

\section{Introduction}
\begin{figure*}[]
    \centering
    \vspace{-1.5mm}
    \includegraphics[width=0.85\linewidth]{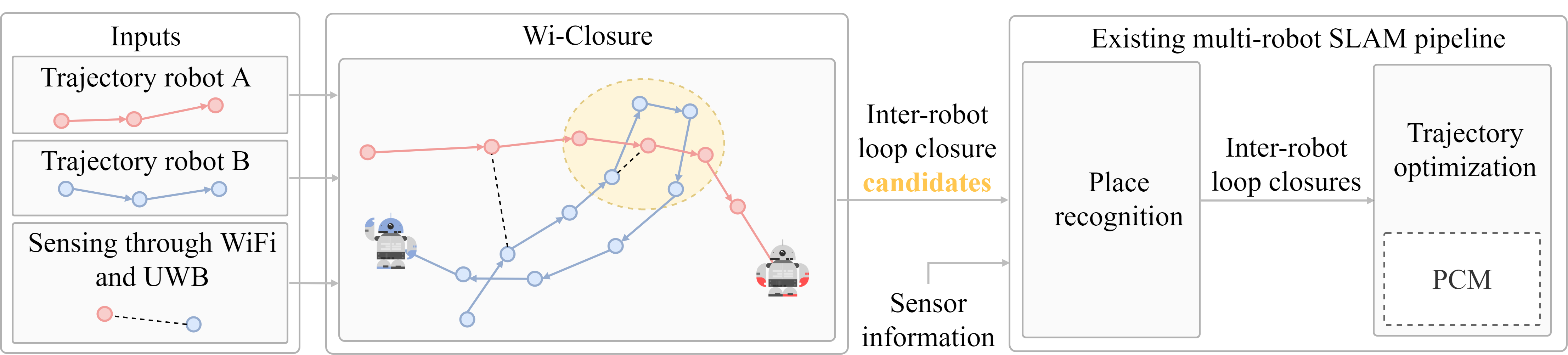}
    \caption{\textit{Wi-Closure} efficiently finds locations where robots' trajectories overlap, as indicated by the yellow area. Only inter-robot loop closures at these locations need to be processed by the multi-robot SLAM pipeline. This increases robustness against perceptual aliasing and decreases overall computation of the pipeline. }
    \label{fig:intro_overview}
\end{figure*}
Loop closure detection has been widely studied as a fundamental aspect of Simultaneous Localization and Mapping (SLAM) \cite{angeli2008fast,hess2016real}. The location estimate of the robot drifts over time due to the noise in the on-board odometer and loop closure detection is essential to correct for this drift by recognizing previously visited places. Without such corrections, the world as perceived by the robot may diverge substantially from reality. %Consequently, loop closure detection is a vital tool to obtain consistent map and location estimates such that these accumulating errors do not deter the robot from completing its task, especially in GPS-denied environments.
Similarly, if multiple robots intend to collaborate, they require a shared situational awareness being consistent with reality, as obtained by multi-robot SLAM. The key to obtaining this shared understanding are inter-robot loop closures. Where regular loop closures constrain the positions of one robot itself, the inter-robot loop closure defines spatial relations between pairs of robots. These inter-robot loop closures enable robots to merge local sensor data into a shared model of the world and obtain relative locations. \\
A common method that robots use to find inter-robot loop closures is place recognition. However, place recognition remains challenging in practice, especially when the environment has repetitive elements \cite{bowman2017probabilistic} and when communication between robots is intermittent. We introduce \textit{Wi-Closure} to address two persistent problems in this setting. %, which uses additional information from the communication itself to find inter-robot loop closures. 
First, since robots do not know each other's location, they may mismatch similar-looking scenes that they have encountered in different locations -- a problem also referred to as perceptual aliasing \cite{angeli2008fast}. Second, during the short intervals that communication between robots is established, feeding a large set of inter-robot loop closures into the multi-robot SLAM pipeline puts a large strain on computational resources \cite{dube2017online}. Repetitive elements further increase computation by falsely recognizing more inter-robot loop closures. %In single-robot SLAM, the robot can use odometry data to get a sense of whether the similar scenes are in approximately the same area. However, this odometry backbone does not connect trajectories of different robots in multi-robot SLAM, making it impossible to readily extend this spatial check to multiple robots. 
Previous work introduced pairwise consistency maximization (PCM) to prevent scene mismatching by identifying false inter-robot loop closures \cite{Mangelson2018}. However, recent research demonstrates that if repetitive elements are present, catastrophic failure of the SLAM algorithm can occur even if using PCM \cite{Ikram2022}. A more robust solution to perceptual aliasing is tracking all possible (mis)matches, resulting in various hypotheses of what the world looks like \cite{Kaess2019}. Unfortunately, working with multiple hypotheses is costly since multiple-hypothesis tracking and planning are computationally complex \cite{Shienman2022}. This makes these methods less viable for real-time execution on commonly available robot hardware.  % which substantially worsens the second problem that \textit{Wi-Closure} aims to address. %Reaching real-time execution then becomes difficult on commonly available hardware. 
% Additionally, methods such as PCM are needed to identify false inter-robot loop closures, which require high additional computation \cite{Computation_PCM}.

%%%%%%%% How does the state of the art approach this challenge? %%%%%%%%%%%
% Recent work appears to have resolved perceptual aliasing in repetitive environments by tracking all (mis)matches with multiple hypotheses in parallel [REFs]. Unfortunately, each added hypothesis comes at great computational cost. Therefore, it may not be viable for real-time execution of multi-robot SLAM on commonly available robot hardware. 

% \textcolor{red}{[Comment] We probably should be more specific here about the set-up of our approach and what setting we consider. Also, we probably first need to finalize our approach on how to solve the multipath problem before we can fully write the following sections. } 
Our approach \textit{Wi-Closure} is a computationally lightweight method that robustly finds inter-robot loop closures in perceptually aliased environments. We use spatial information from WiFi and ultra-wideband (UWB) communication signals to identify where robots' trajectories are close. WiFi is an electromagnetic wave, and thus the receiving robot can locally derive the direction or Angle of Arrival (AOA) to the transmitting robot from the phase information \cite{Kumar2014AccurateIL}. Similarly, commercial UWB devices measure time-of-flight to estimate distance. Importantly, sensing through the communication signal has wide applicability in this setting since it passes through obstacles and thus works in non line-of-sight situations \cite{Jadhav2021}, and it doesn't require the robots to identify each other through vision-based methods, e.g. using Apriltags \cite{Olson2011tags}. % leverages the robustness against perceptual aliasing that multiple hypotheses tracking offers, but reduces computation by finding a smaller set of hypotheses. 
Our method solely uses spatial information and thus it can work seamlessly together with existing place recognition methods based on appearance information. As depicted in Fig.~\ref{fig:intro_overview}, the \textit{Wi-Closure} algorithm is used at the start of the multi-robot SLAM pipeline.

In order to achieve good performance, \textit{Wi-Closure} must also address a major challenge to sensing over the communication signal; namely, it must address multipath propagation of the wireless signal. Multipath refers to the phenomenon where the signal bounces off of various objects to arrive at the receiver from different angles. Consequently, the AOA measurement may include multiple directions, of which at most one is the direct-line path to the other robot. We address this issue with PCM, since only the true direct paths will give consistent pairs of AOA measurements over time. In our hardware experiments, after collecting 4 AOA measurements with in total 3 direct paths and 17 multipaths, we are able to accurately distinguish all direct paths from the multipaths.  %tracking multiple hypotheses, of which the likelihoods are iteratively updated by checking whether a selection of the proposed inter-robot loop closures actually result in a match when additionally using place recognition methods.

Our numerical and hardware experiment results demonstrate that our method efficiently prunes the search space of loop closure candidates by \textbf{99\%} in simulation and \textbf{78.7\%} in hardware experiments, and increases robustness against perceptual aliasing by rejecting up front inter-robot loop closures between distinct places and reducing absolute trajectory estimation error by \textbf{99\%} in simulation and \textbf{89.2\%} in hardware results.\\ 
We summarize the contributions of this paper as follows:
\begin{enumerate}
    \item We introduce a resource efficient approach, \textit{Wi-Closure}, to detect inter-robot loop closures in perceptually aliased environments, based on spatial information from the communication signal. It can work in tandem with existing place recognition methods.
    \item We address the challenging situation of multipath propagation of the communication signal with PCM.
    \item We demonstrate the merits of our approach in terms of robustness against false inter-robot loop closures and improved computation time in simulation with the KITTI dataset and in hardware experiments.
\end{enumerate}

\section{Related Work}
% Add stuff about computation? And sampling method being less robust? Maybe emphasize the computation vs. robustness dilemma in the current literature more
For decades, the majority of research on loop closure detection has focused on a single robot \cite{GalvezLopez2012,Kin2006}. Recently however, loop closure detection algorithms are being adapted to fleets of robots, to ensure reliable and efficient retrieval of shared map and location estimates \cite{Giamou2018,Mangelson2018}. We leverage previous work on sensing over the communication signal to simultaneously address two open problems: 1) computation to match large trajectories is high, and 2) place recognition easily mismatches trajectories in repetitive environments. % These methods face two additional challenges in the multi-robot setting, 1) the odometry backbone does not provide any spatial information on the relative position of scenes visited by different robots, and 2) the methods need to be computationally efficient enough to enable real-time collaboration. 
% Map creation using a single robot has been extensively studied and optimized. However, the area of multi-robot systems for map creation using SLAM is relatively underdeveloped. In single-robot SLAM, odometry measurements are often corrected and used to provide a general map. 
%Furthermore, with the use of multiple robots, accurate loop closure detection becomes increasingly important since robots must know when they have already seen a landmark. 
% \cite{Dennison,DeutschIsaac2016Affm}.

\textbf{Wireless sensing} Extensive research has shown that we can obtain spatial information from wireless signals \cite{Kumar2014AccurateIL,song2019uwb}. Many works use UWB sensors to obtain ranging information between two robots by measuring the time-of-flight of the ultra-wideband signal.\cite{uwb_multi1,uwb_multi2} use the ranging information amongst robots to improve the joint position estimate even without being in line of sight of each other. Recently, \cite{Jadhav2021} also introduced sensing direction from the WiFi communication signal to the robotics community, requiring only a single WiFi antenna and movement of the robot. These innovations avoid the need of bulky equipment and anchors as used in classical works to estimate position, which come with additional infrastructure requirements \cite{arraywifi}. %Instead of using conventional signal strength \cite{signal_strength1,2,3}, it utilizes the motion of robots and attached wifi antennas to emulate antenna arrays in the air to extract bearings without additional infrastructures and bulky equipments. 
% also provide a vision for collaborative robotics to obtain relative spatial information like ranging without relying on the drifting odometers. 
% Source of information rarely incorporated in collaborative robotics. AOA from phase. Distance from ToF or (less accurate) signal strength.\\
% However, the WiFi signal can scatter off various objects, introducing indirect paths between sender and transmitter. This is often referred to as the multipath propagation problem \cite{Vanderveen1998}. Consequently, the AOA measurement can include multiple possible directions, of which maximally one is the true direct path between the two robots. To properly reason about this multipath problem, we can leverage the same frameworks to represent multiple hypotheses as done with repetitive scenes. 

\textbf{Range-only SLAM} Previously, \cite{rangeonly} used UWB sensors in a multi-robot SLAM setting coined range-only SLAM, where distance measurements are directly used as inter-robot loop closures. This avoids the problem of perceptual aliasing, but it only introduces connections between the maps of the robots where the robots are communicating. In realistic scenarios the communication is intermittent, and trajectories can overlap in places where communication is unavailable and where the position estimate is uncertain due to odometer drift. Additional place recognition increases the accuracy of the map by matching these overlapping locations. To our knowledge, we are the first to speed up place recognition using ranging and direction information from the communication signal. 

\textbf{Computation in loop closure} Researchers sought to reduce computation of loop closure detection, e.g. with easily obtainable ORB features for vision-based approaches \cite{ORB}, and efficient look-up trees to match scenes \cite{GalvezLopez2012}. Unfortunately, these methods may result in mismatched maps in perceptually aliased environments \cite{Ikram2022}. %Additionally, incremental versions exist of PCM \cite{Huang2022} and the optimization procedure to  (iSAM2) \cite{Kaess2012}. 
In \cite{Tian2021b} the authors consider sampling a subset of most informative inter-robot loop closures to reduce overall time consumption. However, the authors also note that the performance guarantee of their sampling method decreases if a scene can be potentially matched to many others - i.e. when there is substantial perceptual aliasing. 

\textbf{Perceptual aliasing} Although repetitive scenes are pervasive in many environments, classical place recognition approaches find it notoriously difficult to deal with them. Researchers have focused on simultaneously representing all possible matches as multiple hypotheses in one framework \cite{Hsiao2019}. However, to properly use these multiple hypotheses to determine the best course of action for the robot, we need computationally expensive methods such as data-association belief space planning (DA-BSP) \cite{Pathak2016,Shienman2022}. In DA-BSP the computation time scales exponentially with the hypotheses.\\
We observe that many methods have a trade-off between robustness against perceptual aliasing and computation: increased robustness requires large computation, while computationally efficient methods decrease robustness or perform worse in repetitive environments. Our approach instead aims to improve both computation and robustness against perceptually aliasing. By sensing lightweight information over the communication signal, we efficiently pinpoint where inter-robot loop closures connect scenes that are likely in the same location.

\vspace{-4.5mm}
\section{problem formulation}
\vspace{-1mm}
% Set gamma: with elements alpha, beta
% Add centralized/decentralized, what is communicated
Consider a team of robots operating in an unknown environment, while unaware of their relative positions to each other. All robots obtain odometry measurements to estimate their trajectories locally. These trajectories are spatially connected through measurements on relative position and orientation of the robots, retrieved by sensing over an intermittent communication signal. Based on the information collected so far, we aim to determine where their trajectories overlap with each other, such that these inter-robot loop closure candidates can be further refined by existing place recognition systems.

We consider a classical graph-SLAM setup of a team of robots denoted by the set $\Omega$. Let two robots be represented by $\alpha,\beta\in \Omega$. Each robot estimates its own trajectory $\mathcal{T}^\alpha$ with respect to its local frame $\alpha$. A trajectory is defined by a set of homogeneous transformations from time $t=0$ to $\tau$, as denoted by $\mathcal{T}^\alpha=\{T^\alpha_{t}: t=0,\dots,{\tau} \}$. Here, $T^\alpha_{t}$ is an element in the Special Euclidean Lie group $T^\alpha_{t}\in SE(d)$, consisting of a rotation matrix in the Special Orthogonal Lie group $R^\alpha_t\in SO(d)$ and a translation vector $\textbf{x}^\alpha_t\in\mathbb{R}^d$. Throughout this paper, we adopt the convention of denoting the reference frame as superscript, and the target frame as subscript for $T$, $R$ and $\textbf{x}$, i.e. $\textbf{x}^r_s$ denotes the translation of position $s$ with respect to reference $r$. We estimate the trajectory $\mathcal{T}$ using the graph-SLAM framework, by maximum-likelihood estimation (MLE) of likelihood function $\mathcal{L}$ given some set measurements $\mathcal{Z} $\cite{Cadena2016}.  

\begin{align}\label{eq:MLE}
    \hat{\mathcal{T}} = \arg\max_{\mathcal{T}}  \mathcal{L} = \arg\max_{\mathcal{T}} \prod_k f_k(z_k|\mathcal{T})
\end{align}
Here, factors $f_k(z_k|\mathcal{T})$ are conditional probability density functions that encode the probability of observing an odometry measurement $z_k\in\mathcal{Z}$, given the pose information in $\mathcal{T}$. 
%Assuming Gaussian measurement and process noise, and a linear motion model $h:x^+=Ax+Bu$, this results in two Gaussian factors $f_{odom}$ and $f_{motion}$ for odometry measurements and the motion model respectively.
% \begin{align}
%     f_{\alpha,odom}^i(z_i | \textbf{x}_i, \textbf{x}_{i+1}) &= \frac{1}{4\pi^2|\Sigma|^2}e^{-\frac{1}{2}\|z_i - \textbf{x}_{i+1} + \textbf{x}_i \|_{\Sigma_i}}\\
%     f_{\alpha,motion}^i(u_i | \textbf{x}_i, \textbf{x}_{i+1}) &= \frac{1}{4\pi^2|\Sigma|^2}e^{-\frac{1}{2}\|\textbf{x}_{i+1} - A\textbf{x}_{i} + Bu_i \|_{\Sigma_i}}
% \end{align}
% $\{^{\mathcal{F}_\alpha}\textbf{x}^\alpha^{0},\dots,^{\mathcal{F}_\alpha}\textbf{x}^\alpha^{T}\}$. 
% Meanwhile, when traversing the environment, each robot will sense the environment with some sensing modality, e.g. LiDAR or camera, such that each pose $T_\alpha^{t}$ is associated with a measurement of the environment $z_{e,\alpha}^t$. %Robots can exchange these measurements and search for a set $G$ of inter-robot loop closures. 
% In this work, we do not directly address how robots should process this often high-dimensional sensor information to detect inter-robot loop closures, but rather focus on determining which sensor measurements $z_{e,\alpha}^t$ should be exchanged and processed in the first place. For this, we propose to leverage the spatial information gathered from the communication signal. 
\subsection{Multi-robot SLAM with wireless measurements}
We propose to use information from the communication signal between robots to relate their trajectories. UWB and WiFi signals provide measurements on the distance $d$ between robots, and the direction $\phi$ of the signal-transmitting robot with respect to the signal-receiving robot, respectively. % Consider robot $\alpha$ at position $\textbf{x}_p^\alpha$, obtaining measurements on distance $d$ and direction $\phi$ to robot $\beta$ at position $\textbf{x}_k^\beta$. Then, the translation between these positions $\textbf{x}^p_k$ is an element of the shared trajectory estimate $\mathcal{T}^{\alpha,\beta} = \{\mathcal{T}^{\alpha}, \mathcal{T}^{\beta}\}$. 
Previously, \cite{Naseri2019} formulated these AOA and distance measurements as factors to solve a localization problem. We adopt this formulation, with factors $f_{uwb}(d|\mathcal{T}^{\alpha,\beta})$ and $f_{aoa}(\phi|\mathcal{T}^{\alpha,\beta})$ defined as follows. 

\begin{align}\vspace{-3mm}
    % f_{uwb}(d|\mathcal{T}^\alpha,\mathcal{T}^\beta)&= c_{1} \exp{\left(\frac{-1}{\sigma_{\alpha,\beta}} (d-\|\textbf{x}^\alpha_p - R_\beta^\alpha\textbf{x}^\beta_k \|_2)\right)}\\
    % f_{aoa}(\phi|\mathcal{T}^\alpha,\mathcal{T}^\beta)&= c_{2}\exp{\left(\kappa_{\alpha,\beta}\textbf{u}^T(\phi)\frac{\textbf{x}^\alpha_p-R^\alpha_\beta\textbf{x}^\beta_k}{\|\textbf{x}^\alpha_p - R^\alpha_\beta\textbf{x}^\beta_k \|_2}\right)}
    f_{uwb}(d|\mathcal{T}^{\alpha,\beta})&= c_{1} \exp{\left(\frac{-1}{\sigma_{\alpha,\beta}^2} (d-\|\textbf{x}^p_k \|_2)^2\right)}\\
    f_{aoa}(\phi|\mathcal{T}^{\alpha,\beta})&= c_{2}\exp{\left(-\kappa_{\alpha,\beta}\textbf{u}^T(\phi)\frac{\textbf{x}^p_k}{\|\textbf{x}^p_k \|_2}\right)}\label{eq:factor_aoa}
\end{align}

where $c_{1}=\frac{1}{\sqrt{2\pi\sigma_{\alpha,\beta}^2}}$, $c_{2}=\frac{1}{2\pi I_0(\kappa_{\alpha,\beta})}$ and $\textbf{u}=[\cos \phi, \sin \phi]^T$. Here, $I_0(.)$ is the modified Bessel function of the first kind of order zero, $\sigma_{\alpha,\beta}^2$ the variance of the distance measurement, and $\kappa_{\alpha,\beta}$ a concentration parameter computed as the inverse of the AOA variance, i.e. $\kappa_{\alpha,\beta}>10\ \text{rad}^2$. Then, these factors are combined into one factor.
% \begin{gather}
%     \begin{split}
%         \MoveEqLeft
%         f_{comm}^i(d,\phi|\mathcal{T}^{\alpha,\beta})= \\
%         &c_{1}c_{2} \exp{\left(\frac{-1}{\sigma_{\alpha,\beta}^2} (d-\|\textbf{x}^p_k \|_2)^2 - \kappa_{\alpha,\beta}\textbf{u}^T(\phi)\frac{\textbf{x}^p_k}{\|\textbf{x}^p_k \|_2}\right)}
%     \end{split}\raisetag{3\baselineskip}
% \end{gather}
\begin{equation}
f_{comm}(d,\phi|\mathcal{T}^{\alpha,\beta})=f_{aoa}(\phi|\mathcal{T}^{\alpha,\beta})f_{uwb}(d|\mathcal{T}^{\alpha,\beta}) \label{eq:factor_comm}
\end{equation}

\begin{figure}
    \centering
    \includegraphics[width=0.7\linewidth]{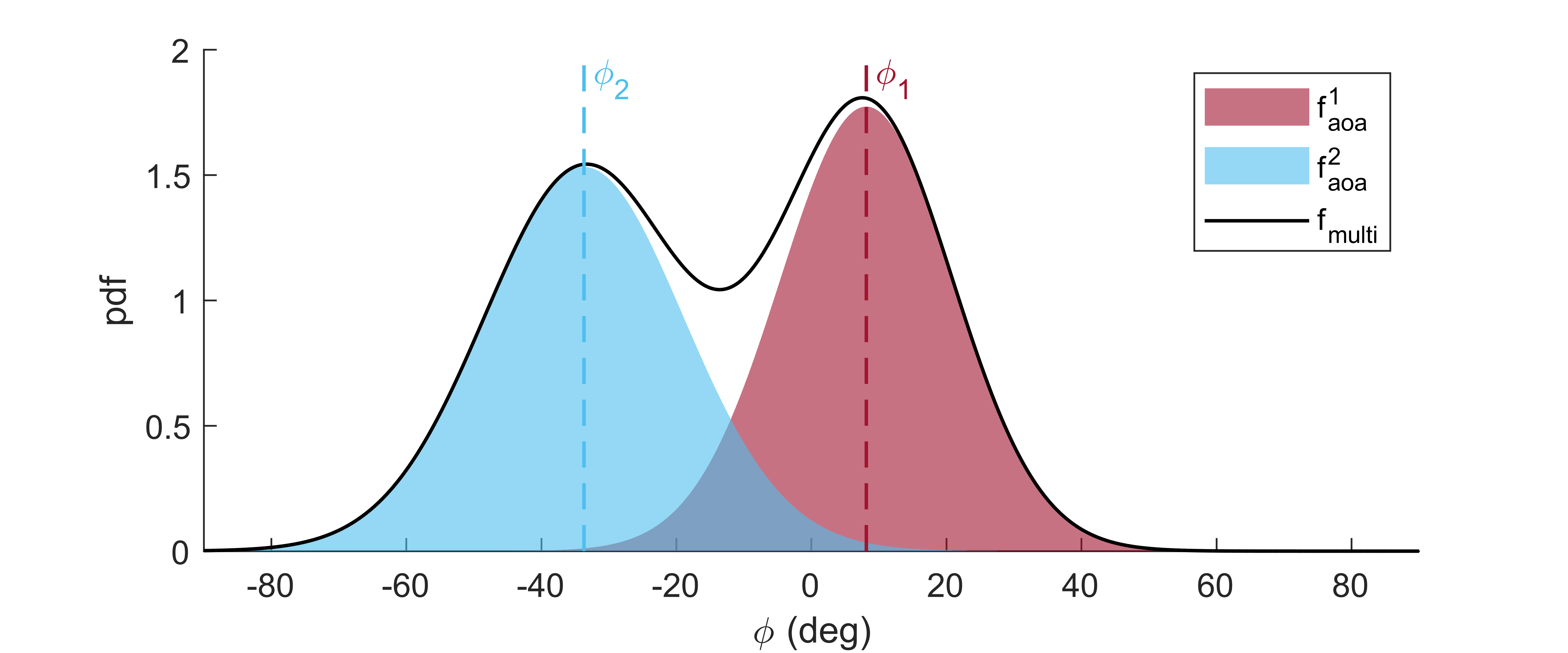}
    \includegraphics[width=0.7\linewidth]{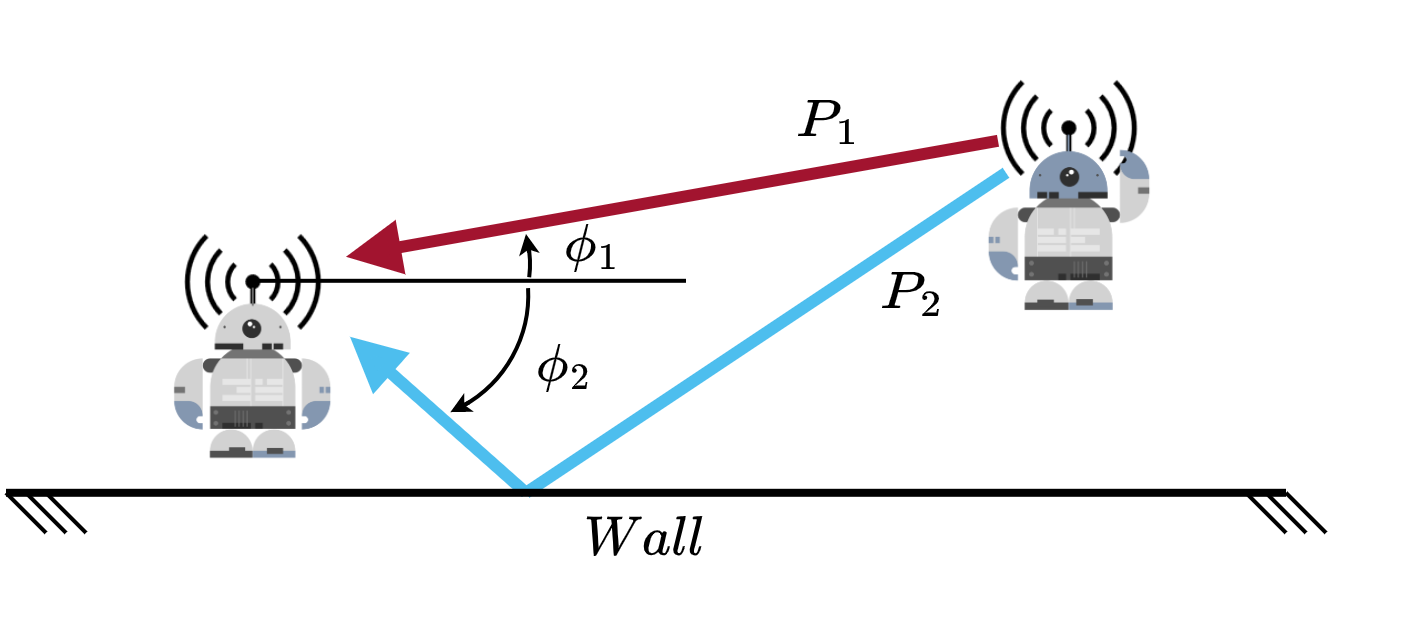}
    \caption{The communication signal can reach the robot through different paths $\mathcal{P}=\{P_1,P_2\}$, resulting in multiple Gaussian modes $f^i_{aoa}$ in the AOA measurement $f_{multi}$.}
    \label{fig:multipath}
\end{figure}

Importantly, \cite{Naseri2019} models the AOA measurement as a single Gaussian. However, this may not realistically represent the AOA measurement in practice due to multipath propagation of the signal. Objects in the environment can reflect the communication signal, causing it to arrive at the robot via different paths as shown in Fig.~\ref{fig:multipath}. These paths cause the AOA measurement to have multiple (approximately Gaussian) modes. For the $j^{th}$ AOA measurement, we parameterize these paths by the set of binary variables $\mathcal{P}^j=\{P_1^j, P_2^j, ...,P_n^j \}$, where $P_i^j=1$ indicates that path $i$ in the $j^{th}$ AOA measurement is the direct path. Then, the multimodal AOA measurement $f_{multi}$ can be modeled as a marginalization over multiple $f_{aoa}$.
\begin{align}
    f_{multi,j}(\phi|\mathcal{T}) &= \sum_i^{|\mathcal{P}|} 
    f^i_{aoa,j} (\phi | \mathcal{T}, P_i^j=1)p(P_i^j=1)
\end{align}
% We now need to solve our problem for $|\mathcal{P}|$ different scenario's, of which only one scenario is assumed to contain the factor $f_{aoa}^i$ associated with the true direct path $P_i=1$. 

% Each mode $i$ results in a different MLE estimate $\hat{\mathcal{T}}^i_{\alpha,\beta}$. We retrieve the MLE estimates of each mode $\hat{\mathcal{T}}^i_{\alpha,\beta}$ by replacing multimodal $f_{multi}(\phi| \mathcal{T}_{\alpha,\beta})$ in equation \ref{eq:MLE} with Gaussian distribution $f_{aoa} (\phi | \mathcal{T}_{\alpha,\beta}, P_i)$ for all $P_i\in\mathcal{P}$. In the multipath propagation case, the objective is then altered to search for the union of all sets of inter-robot loop closures $G_M$.\vspace{-5mm}
% \begin{center}
%     \begin{equation}
%         G_M=\bigcup_{i=1}^{|\mathcal{P}|} G^i
%     \end{equation}
%     $G^i = \{(\textbf{x}^\alpha, \textbf{x}^\beta)_{kp} : p(u_{kp}=1) > p_M, \hat{\mathcal{T}}^i_{\alpha,\beta}\}$\\[6pt] $i=1,\dots,|\mathcal{P}|$
% \end{center}

Each AOA measurement has at most one path as the true direct path. Therefore, an important problem that \textit{Wi-Closure} addresses is how to determine the set of direct paths for multiple AOA measurements, which we denote by realization $\mathcal{R}=\{P_i^j\ |\ P_i^j=1,\ \forall j\ \sum_{i=1}^n P_i^j \leq 1 \}$.
Then we can obtain an estimate of what we will refer to as \emph{\textbf{the shared robot trajectory $\mathcal{T}^{\alpha,\beta}$}}, by adding communication factors $f_{comm,j}^i$ corresponding to $P_i^j\in\mathcal{R}$ to the MLE in Equation \ref{eq:MLE}.

\subsection{Inter-robot loop closures as a set of nearby poses}
With the spatial information contained in ${\mathcal{T}}^{\alpha,\beta}$, we are interested in retrieving the positions where the trajectories of robot $\alpha$ and robot $\beta$ are nearby each other. To assess whether some position $\textbf{x}^\alpha_p\in \mathcal{T}^\alpha$ of robot $\alpha$ is nearby some position $\textbf{x}^\beta_k\in \mathcal{T}^\beta$ of robot $\beta$, we use the Mahalanobis distance.  
\begin{equation}\label{eq:MH_distance}
    d_{MH}(\textbf{x}^\alpha_p, \textbf{x}^\beta_k) = \sqrt{(\textbf{x}^p_k)^\top\  \Sigma_{p,k}^{-1}\ \textbf{x}^p_k}
\end{equation}
The main objective of \textit{Wi-Closure} is then to efficiently find all position-pairs $(\textbf{x}^\alpha_p, \textbf{x}^\beta_k)$ that have a Mahalanobis distance smaller than some threshold $D$, as collected in set $G$ and are thus good loop closure candidates.
\begin{equation}\label{eq:G}
    G = \{(\textbf{x}^\alpha_p, \textbf{x}^\beta_k)\ |\ d_{MH}(\textbf{x}^\alpha_p, \textbf{x}^\beta_k) < D,\ \mathcal{R} \}
\end{equation}
Note that a different set $G$ will be found for different guesses of the direct paths $P^j_i = 1$, i.e. for different realizations $\mathcal{R}$.

\section{Approach}
\begin{figure}[t]
    \centering
    \includegraphics[width=\linewidth]{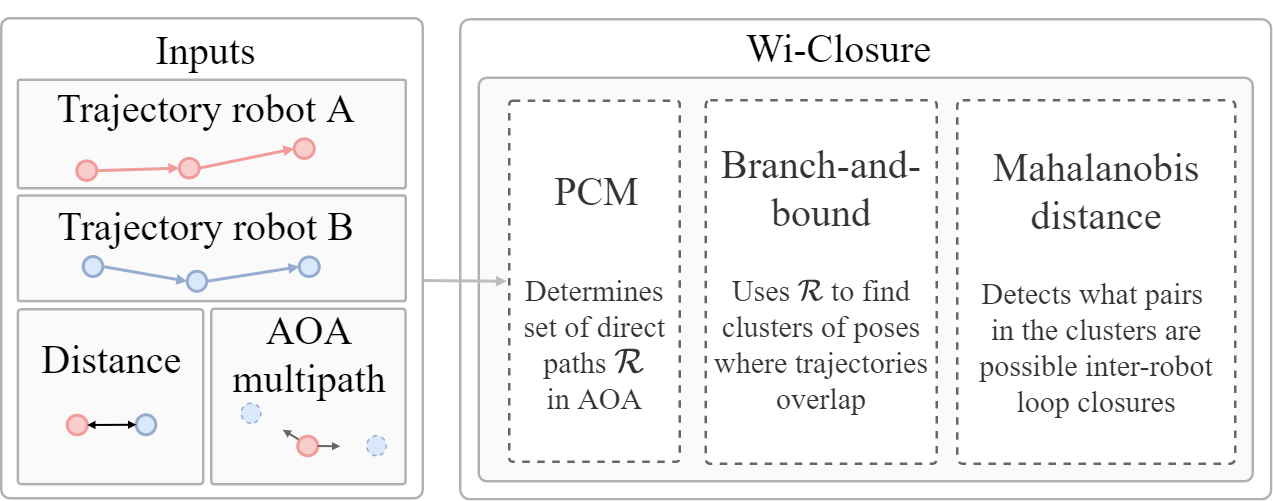}
    \caption{Overview of \textit{Wi-Closure}.}
    \label{fig:approach_overview}
\end{figure}
This section explains the approach taken by \textit{Wi-Closure} on three core aspects, of which an overview is shown in Fig.~\ref{fig:approach_overview}. First, we show how trajectory information and communication measurements combined can give a reliable estimate of the shared trajectory. Here, we reject spurious multipaths in the communication measurements using PCM. Secondly, an algorithm akin the branch-and-bound algorithm quickly finds areas where the trajectories overlap. Lastly, for each position pair in the overlapping areas, we determine whether it is a candidate inter-robot loop closure using the Mahalanobis distance.  
\subsection{A shared trajectory estimate while rejecting AOA multipath}
\begin{figure}[b!]
    \centering
    \vspace{-2mm}
    \includegraphics[width=0.9\linewidth]{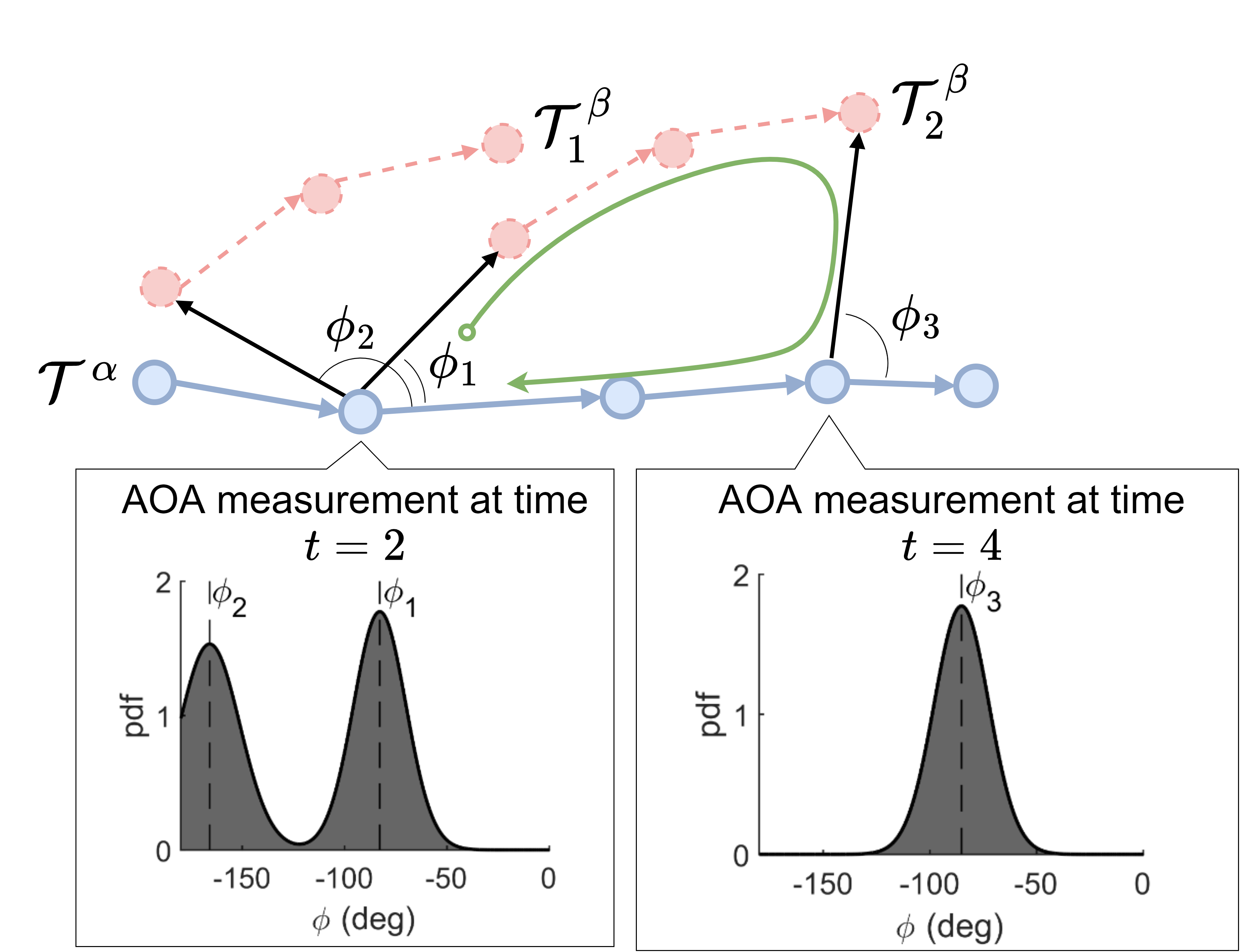}
    \caption{From robot $\alpha$'s perspective, at time $t=2$ the trajectory of robot $\beta$ could be either at $\mathcal{T}^\beta_1$ or $\mathcal{T}^\beta_2$ due to the multipath in the AOA measurement (black arrows). By additionally using the AOA measurement at time $t=4$, PCM determines that the paths corresponding to $\phi_1$ and $\phi_3$ are direct paths, since they can form a loop (green arrow). Therefore, $\mathcal{T}^\beta_2$ is robot $\beta$'s real trajectory.}
    \vspace{-5mm}
    \label{fig:PCM}
\end{figure}
\textit{Wi-Closure} uses as input the robot trajectories $\mathcal{T}^\alpha$ and $\mathcal{T}^\beta$ and communication factors $f_{uwb}$ and $f_{multi}$. Due to the multipath propagation problem, the AOA measurement determining factor $f_{multi}$ can be multimodal, while only one mode possibly gives useful information on the direct path. In this section we therefore show how PCM finds a set of direct paths, denoted by realization $\mathcal{R}$. Communication factors $f_{comm}$ are then constructed using the paths in $\mathcal{R}$ to connect the robot trajectories while avoiding AOA multipaths.

The PCM method first determines for each pair of measurements whether they are consistent with each other \cite{Mangelson2018}. As shown in Fig.~ \ref{fig:PCM}, two communication measurements are consistent with each other if we can traverse them and the odometry backbone of the robot trajectories in a loop. Let $T^i_j$ and $T^k_l$ be two transformations defined by some communication factor $f_{comm,1}$ and $f_{comm,2}$ respectively, and define the trajectory sections $T^j_k = (T^\alpha_j)^{-1}(T^\alpha_k)$ and $T^l_i = (T^\beta_l)^{-1}(T^\beta_i)$. Then, if the loop is closed the following equality should hold.
\begin{equation}
	T_{loop} = T^i_j\ T^j_k\ T^k_l\ T^l_i = I
\end{equation}
To account for noise in the transformation estimates, we identify consistent loops using the Mahalanobis distance $d_{PCM}$. For this we use Lie algebra to express the transformation as a 6D vector $\xi_{loop}\in \mathfrak{se}(3)$ with $\xi_{loop}=\log (T_{loop})$.
\begin{equation}
    d_{PCM} = \sqrt{\xi_{loop}^\top \Sigma^{-1}_{loop} \xi_{loop}}
\end{equation}
where $ \Sigma_{loop}$ is the covariance matrix corresponding to $\xi_{loop}$. Then, the largest set of communication measurements that are all consistent with each other, gives us a set with likely only measurements of direct paths. Hence we have found realization $\mathcal{R}$ with which we can estimate how the trajectories $\mathcal{T}^\alpha$ and $\mathcal{T}^\beta$ are positioned with respect to each other.

\subsection{Efficiently finding trajectory overlap}
\begin{figure*}[t]
    \centering
    \includegraphics[width=0.8\linewidth]{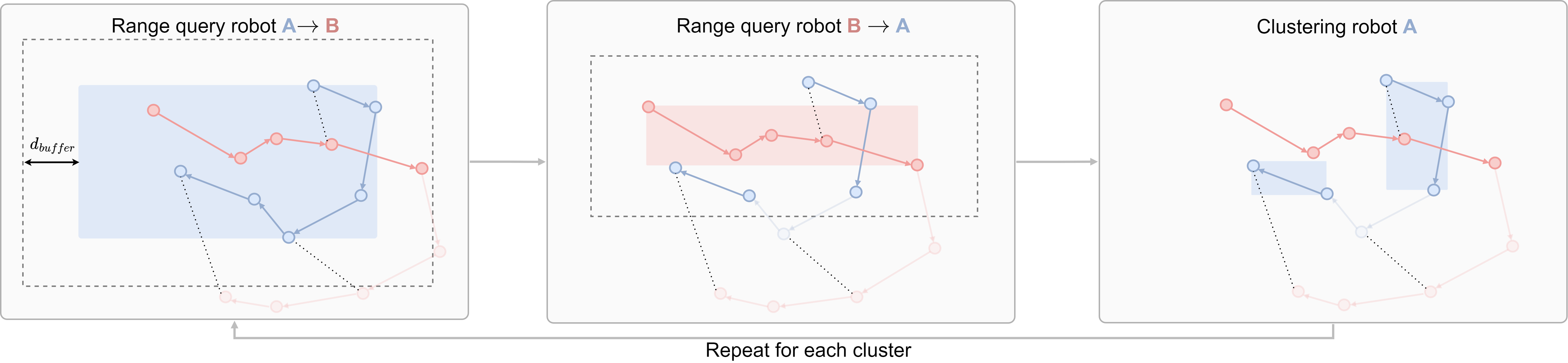}
    \caption{Finding area's where trajectories overlap by iterative refinement of overlapping bounding boxes.}\vspace{-5mm}
    \belowcaptionskip=1pt % below caption
    \label{fig:trajectory_overlap_finding}
\end{figure*}
% Computing for all possible position pairs $\textbf{x}^\alpha_p\in\mathcal{T}^\alpha$ and $\textbf{x}^\beta_k\in\mathcal{T}^\beta$ whether they should be included in set $G$ would be a computationally expensive task. Hence, we propose to do this only for position pairs within some smaller clusters. 
We quickly find clusters where trajectories overlap using a method similar to the classical branch-and-bound algorithm. As shown in Fig.\ref{fig:trajectory_overlap_finding}, our approach first bounds the area's traversed by robots $\alpha$ and $\beta$, and selects the poses within this overlap. These poses are divided into smaller clusters, and the process is repeated for each cluster. 
The initial bounds on the area are found by selecting the minimum and maximum position coordinate along each dimension. However, we need to account for possible distance between true loop closures and uncertainty in the poses. We add $d_{buffer}$ to the bounds, which is computed such that we retain all position pairs that are later included as inter-robot loop closures when computing the Mahalanobis distance.
\begin{equation}
    d_{buffer} = D \sigma_{UB} + R_{sensor}
\end{equation}
where $D$ is the threshold of the Mahalanobis distance used in Equation \ref{eq:G}, $\sigma_{UB}$ is an upper bound to the worst-case uncertainty that we can expect in any direction for any position pair, and $R_{sensor}$ is the range of the sensor that will determine at what distance we can expect to find loop closures.

First, consider the maximum uncertainty $\sigma_{kp}^{max}$ for the translation $\textbf{x}^k_p$ between a single position pair, computed as the square root of the largest eigenvalue of $\Sigma_{kp}$. We then aim to distributively find $\sigma_{UB}$ that is an upper bound to $\sigma_{kp}^{max}$ for any two poses $\textbf{x}^\alpha_{p}\in\mathcal{T}^{\alpha}$ and $\textbf{x}^\beta_{k}\in\mathcal{T}^{\beta}$.
\begin{align}
    \sigma_{UB}^2 \geq \max_{k,p}(\sigma_{kp}^{max})^2 = \max_{k,p} \lambda_{max} (\Sigma_{kp}), \quad p,k\in t
\end{align}
where $\lambda_{max}$ is the largest eigenvalue of $\Sigma_{kp}$. Secondly, $\textbf{x}^p_k$ is rewritten as a pose composition of poses in the local frames.
\begin{align}
    \textbf{x}^p_k% = \textbf{x}^\alpha_p - T^\alpha_\beta\textbf{x}^\beta_k
    = \ominus \textbf{x}^\alpha_p  \oplus \textbf{x}^\alpha_\beta \oplus \textbf{x}^\beta_k
\end{align}
This allows us to determine an upper bound on $\lambda_{max}(\Sigma_{kp})$.
\begin{align*}
    \lambda_{max}(\Sigma_{kp}) &\approx \lambda_{max}(\Sigma_{\alpha p} + J_{\alpha\beta}^T \Sigma_{\alpha\beta}J_{\alpha\beta} + J_{\beta k}^T \Sigma_{\beta k}J_{\beta k})\\ %&\leq \lambda_{max}(\Sigma_{kp}^{RH}) \\
%     &= \lambda_{max}(\Sigma_{\alpha p} + J_{\alpha\beta}^T \Sigma_{\alpha\beta}J_{\alpha\beta} + J_{\beta k}^T \Sigma_{\beta k}J_{\beta k})
% \end{align*}
% Then, we can further split this term.
% \begin{align*}
%     \lambda_{max}(\Sigma_{kp}^{RH}) %&= \lambda_{max}(\Sigma_{\alpha p} + J_{\alpha\beta}^T \Sigma_{\alpha\beta}J_{\alpha\beta} + J_{\beta k}^T \Sigma_{\beta k}J_{\beta k}) \\
    &\leq \lambda_{max}(\Sigma_{\alpha p}) + \lambda_{max}(J_{\alpha\beta}^T \Sigma_{\alpha\beta}J_{\alpha\beta}) \\&\quad + \lambda_{max}(J_{\beta k}^T \Sigma_{\beta k}J_{\beta k})
    = \lambda_{kp}^{UB}
\end{align*}
where $J$ is the Jacobian of $\textbf{x}$. Then, worst-case uncertainty $\sigma_{UB}$ is distributively computed as the maximum of all variances $\lambda_{kp}^{UB}$ between any position pair $\textbf{x}^\alpha_{p}\in\mathcal{T}^{\alpha}$ and $\textbf{x}^\beta_{k}\in\mathcal{T}^{\beta}$.
\begin{align*}
    \sigma_{UB}^2 &= \max_{k,p}( \lambda_{kp}^{UB} )%\\
    % &= \max_{p}( \lambda_{max}(\Sigma_{\alpha p})) + \lambda_{max}(J_{\alpha\beta}^T \Sigma_{\alpha\beta}J_{\alpha\beta}) \\&\quad + \max_{k}(\lambda_{max}(J_{\beta k}^T \Sigma_{\beta k}J_{\beta k}) ))\\
    \geq \max_{k,p}( \lambda_{max} (\Sigma_{kp})) = \sigma_{max}^2
\end{align*}
Note that $\Sigma_{\alpha\beta}$ is computed using the communication factor $f_{comm}$ and can be taken out of the maximization. We need to take the maximum of largest eigenvalues only of covariance matrices $\Sigma_{\alpha p}$ and $\Sigma_{\beta k}$, which are both computed distributively from the MLE trajectory estimates $\mathcal{T}^\alpha$ and $\mathcal{T}^\beta$. The graph-SLAM formulation using factors $f(z|x)$ enables us to retrieve these covariance matrices with a Gaussian approximation.
\begin{equation}\label{eq:uncertainty_from_factors}
	\Sigma = \left( -E_z\left[ \frac{\partial^2 \log f(\textbf{z}|\textbf{x})}{\partial \textbf{x}^2} \bigg| \textbf{x} \right] \right)^{-1}
\end{equation}

\subsection{Identifying inter-robot loop closures}
A position pair $(\textbf{x}^\alpha_p, \textbf{x}_\beta^k)$ identified by the clustering in the previous section is included into set $G$ as a candidate inter-robot loop closure if the Mahalanobis distance is smaller than $D$ (Equation \ref{eq:G}). 
% To compute this, we use pose and uncertainty information from the robot trajectories and the communication measurements, where we assume in this section that the chosen realization $\mathcal{R}$ of communication measurements only include direct paths and no multipaths. 
This requires an estimate of the relative distance and corresponding uncertainty between these two poses, which we extract from our MLE to the shared trajectory estimate. When solving for this MLE, we could include all communication factors corresponding to $P_i^j\in\mathcal{R}$ simultaneously into our optimization problem. However, when connecting trajectories $\mathcal{T}^\alpha$ and $\mathcal{T}^\beta$ through multiple communication factors this is a nonlinear optimization, which also alters the solution to the local trajectory estimates $\mathcal{T}^\alpha$ and $\mathcal{T}^\beta$. Meanwhile, a single communication measurement has a straightforward solution, since this constraint only re-positions the trajectories with respect to each other and does not alter the local solutions to $\mathcal{T}^\alpha$ and $\mathcal{T}^\beta$. For each position pair $(\textbf{x}^\alpha_p, \textbf{x}^\beta_k)$ we choose one communication measurement connecting the trajectories at positions $\textbf{x}^\alpha_{c1}$ and $\textbf{x}^\beta_{c2}$. Then, pose and uncertainty information is propagated from $\textbf{x}^\alpha_p$ to $\textbf{x}^\beta_k$.
\begin{align}
	T^p_k &= T^p_{c1} T^{c1}_{c2} T^{c2}_{p} \\
	\Sigma_{pk} &= \Sigma_{p c1} + J_{c1 c2}^\top \Sigma_{c1 c2} J_{c1 c2} + J_{c2 k}^\top \Sigma_{c2 k} J_{c2 k} \label{eq:uncertainty_prop}
\end{align}
where $J_{ij}$ is the Jacobian of $T^i_j$. 

For each position pair, our algorithm uses the communication link that results in minimum route length from $\textbf{x}^\alpha_p$ to $\textbf{x}^\beta_k$ over the odometry backbone and communication link. The subsequently retrieved values for $T^p_k$ and $\Sigma_{pk}$ (using equation \ref{eq:uncertainty_from_factors} and \ref{eq:uncertainty_prop}) are used to compute the Mahalanobis distance $d_{MH}(\textbf{x}^\alpha_p, \textbf{x}^\beta_k)$, which determines whether the position pair should be included in set $G$.

\section{Experiments}
In this section, We evaluate \textit{Wi-Closure} through simulation and hardware experiments. Our results show that \textit{Wi-Closure} can efficiently and robustly detect loop closures, while processing large trajectories in batches and in repetitive environments. Our approach also successfully handles the multipath phenomenon of the wireless signal in practice. %First, we show the performance of our approach using the KITTI dataset by simulating the wireless measurements using noise model with practical noise parameters upon our previous work \cite{ijrr}. In section , we present the the evaluation in a \textcolor{red}{square meter } unfinished shell space that has highly repetitive features. By collecting AOA and ranging measurements, we show our approach can 

% List of intended simulation results on the KITTI dataset:
% \begin{enumerate}
% \item Picture using only the original DiSCO-SLAM approach vs. adding Wi-Closure. The parameters are tuned such that map-merging fails in the first case, but it still works when adding Wi-Closure. These pictures are supported by numbers on Average Trajectory Error (compared to groundtruth with GPS data)
% \item For the same situation: compare computation time of the whole pipeline when using only the original pipeline vs. when adding Wi-Closure.
% \item Simulation of multipath: show metrics how often PCM still includes multipath and what fraction of true direct paths it rejects (i.e. a confusion table)
% \end{enumerate}
% Background: experiment site, baseline method, how do we perform our experiments
\begin{figure}[b!]
    \centering
    \includegraphics[width=0.8\linewidth]{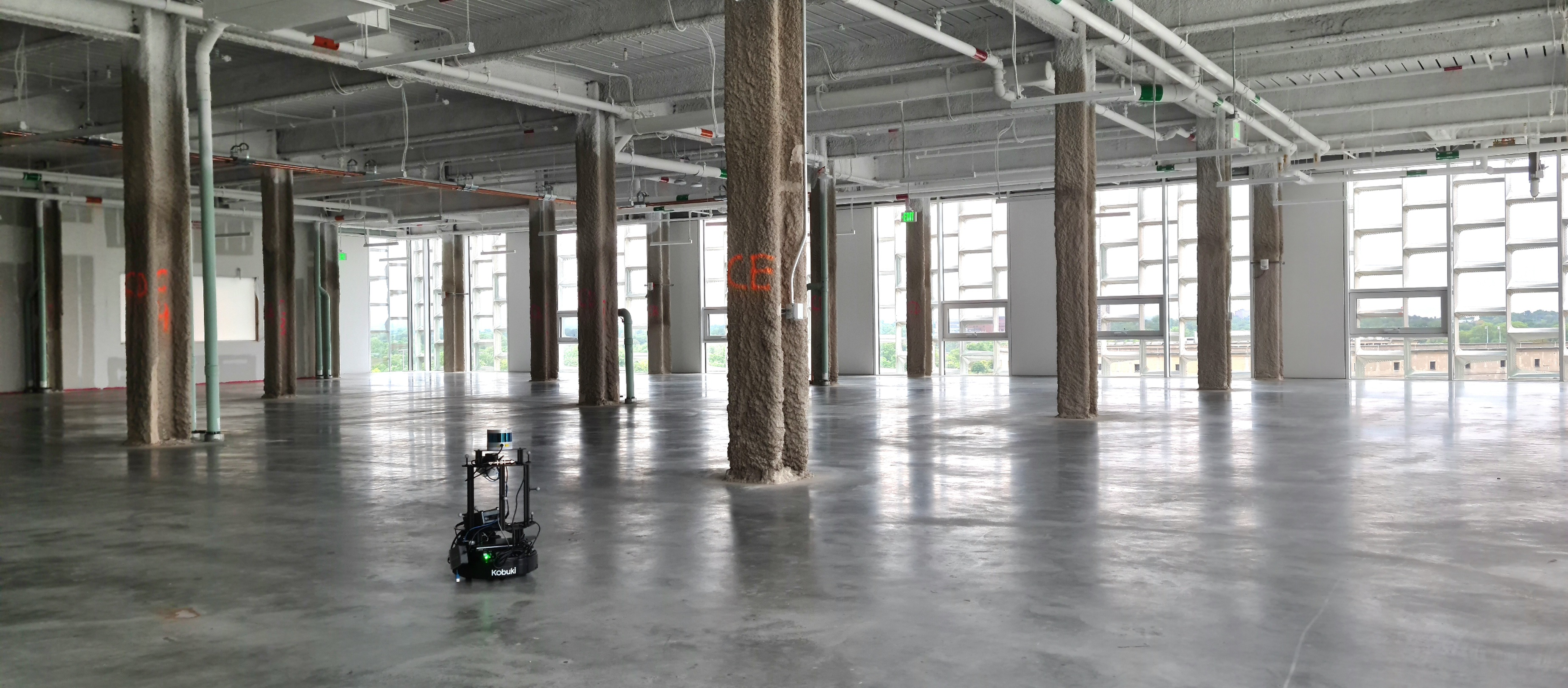}
    \caption{The $25 m \times 23 m$ testing field for hardware experiments with highly repetitive features including identical pillars. }
    \label{fig:shellspace}
\end{figure}

\subsection{Simulation experiments}
Simulations are performed on the KITTI 08 dataset modified by \cite{Huang2022}, where a trajectory is split into sections to emulate the multiple robot case with trajectory overlap. Since this dataset does not contain measurements from the wireless signal, we simulate these based on the groundtruth (GPS) trajectory. We use a standard deviation of $0.5\ m^2$ for distance and $10 \deg$ for AOA, based on previous work characterizing these measurements \cite{Jadhav2021}. %The KITTI 08 dataset 100Hz raw inertial measurement unit (IMU) data is recorded with suitable temporal consistency to be used with LIO-SAM \cite{Shan2020}. The modified version is a synthetic two-robot dataset where time-stamps have been adjusted to incorporate overlap and rendezvous. 
All comparisons are performed %using playback of previously gathered data, in the form of a rosbag file, 
on a desktop computer running an Intel i9 5.2GHz processor in Ubuntu Linux 18.04.
We assess the efficacy of \textit{Wi-Closure} by comparing the performance of the multi-robot DiSCo-SLAM pipeline with and without using \textit{Wi-Closure}. The performance is assessed based on average trajectory error (ATE) and the number of correctly and falsely included inter-robot loop closures. To determine which loop closures are true and false, we use a GPS-based groundtruth trajectory and define true inter-robot loop closures as positions that are at a maximum distance of $35\ m$, such that the LiDAR scans with a range of $30\ m$ overlap for $20\%$.  

Originally, \cite{Huang2022} tuned the parameters of the DiSCO-SLAM algorithm such that it has good performance against mismatching on the modified KITTI 08 dataset. However, we argue that parameters do not necessarily generalize to other environments (as we show in our hardware experiments). We therefore consider a worse set of parameters in this comparison. Then, we show that while the original DiSCO-SLAM pipeline fails with this parameter set, using the same set of parameters and adding Wi-Closure can still recover good performance.
%for multi-robot SLAM for data-efficient exchange of LiDAR observations between robots. The algorithm includes a two-stage global and local optimization framework for distributed multi-robot SLAM. \textit{WiClosure} uses DiSCo-SLAM's framework and builds on top of it by adding 

\begin{figure}[]
\centering
% \begin{subfigure}{.53\linewidth}
%   \centering
  \includegraphics[width=0.5\linewidth]{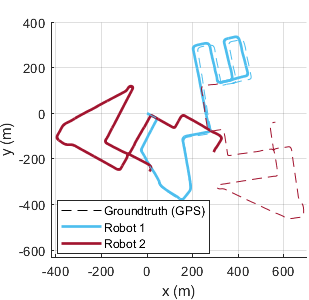}
%   \caption{The optimized trajectory from Disco-SLAM without using \textit{Wi-Closure}.}
%   \label{fig:sub1}
% \end{subfigure} \hfill%
% \begin{subfigure}{.42\linewidth}
%   \centering
  \includegraphics[width=0.42\linewidth]{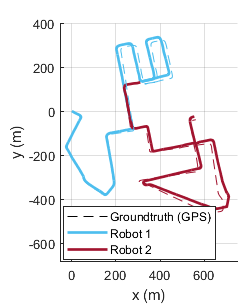}
  % \caption{The optimized trajectory from Disco-SLAM using \textit{Wi-Closure}.}
  % \label{fig:sub2}
% \end{subfigure}
\caption{Simulation results in KITTI 08 dataset. Left: optimized trajectory from Disco-SLAM without \textit{Wi-Closure}. Right: optimized trajectory from Disco-SLAM using \textit{Wi-Closure}.}
\label{fig:simulation_results}
\end{figure}

Table \ref{tab:sim_results} shows that including \textit{Wi-Closure} in the multi-robot SLAM pipeline results in a lower ATE. Fig.~\ref{fig:simulation_results} shows that the baseline approach includes too many false loop closures resulting in catastrophic failure.
Also, without \textit{Wi-Closure}, DiSCO-SLAM processes all 1,099,101 position pairs as possible loop closures, of which 5544 are true loop closures. Meanwhile, \textit{Wi-Closure} substantially reduces this search space to 7,049 inter-robot loop closures, of which 3,631 are true positive loop closures. This comes at a cost of missing 1,913 potential loop closures.
\begin{table}[b!]\vspace{-1mm}
\centering
\begin{tabular}{ |p{4cm}|p{1.5cm}|p{1.5cm}|}
 \hline
   & \textbf{Baseline} &  \textbf{\textit{Wi-Closure}}\\
 \hline
 ATE (m) & 66.1 & 1.3 \\
 \hline
%  Correctly included true LC & 5544 & 3631 \\
%  \hline
 Correctly rejected false LC (\%) & N/A & 99 \\
 \hline
%  Incorrectly included false LC & 1093557 & 3418 \\
%  \hline
 Missed true LC (\%) & 0 & 1 \\
 \hline
 Total Computation Time (s) & 896 & 411 \\
 \hline
 Total Wi-Closure time (s) & N/A & 53 \\
 \hline
\end{tabular}
\caption{Loop Closure (LC) performance comparison between \textit{Wi-Closure} and DiSCO-SLAM in the KITTI Dataset.}
\label{tab:sim_results}
\end{table}
% \begin{figure}[ht!]
%     \centering
%     \subfloat[\centering The optimized trajectory from Disco-SLAM without using \textit{Wi-Closure}.]{{
%     \includegraphics[width=.5\linewidth]{Figures/Trajectory_NoWiclosure_Simulation.png}
%     }}
%     \subfloat[\centering The optimized trajectory from Disco-SLAM using \textit{Wi-Closure}]{{
%     \includegraphics[width=.5\linewidth]{Figures/Trajectory_Wiclosure_Simulation.png}
%     }}
%     \caption{Simulation results on the KITTI 08 dataset. }
% \end{figure}
% Simulation results show that the Wi-Closure algorithm is robust against mismatching. %We test our algorithm using the KITTI dataset with a modified DiSCo-SLAM algorithm as a baseline. The modification on the DiSCo-SLAM algorithm is intended to highlight Wi-Closure's robustness to failure while the baseline algorithm fails to produce a correct map. 
% compared to the baseline algorithm that detects 5,544 true positive loop closures, the difference is due to the number of candidates \textit{Wi-Closure} is considering in comparison to the baseline method. %The baseline algorithm does not reject any loop closure candidates and evaluates all the candidates while the \textit{Wi-Closure} algorithm filters the candidates. 
%In total, the baseline algorithm accepts 1,099,101 potential loop closure candidates while \textit{Wi-Closure} accepts 7,049 potential loop closure candidates. 
As a result, the whole pipeline takes 896 seconds for the baseline algorithm, whereas adding \textit{Wi-Closure} reduces it to 464 seconds. of which 53 seconds caused by added computation of the \textit{Wi-Closure} module.\\ %This significant reduction in time is greatly increased when considering that the baseline algorithm fails to compute the loop closures in the given time constraint. Although \textit{Wi-Closure} fails to consider 1,913 true loop closures, it correctly includes a large portion of the true loop closures and excludes an even greater number of false loop closures. \\
        \begin{figure}[b!]
            \includegraphics[width=.45\linewidth]{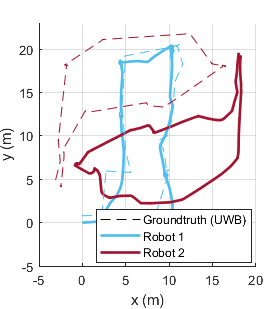}
            \includegraphics[width=.45\linewidth]{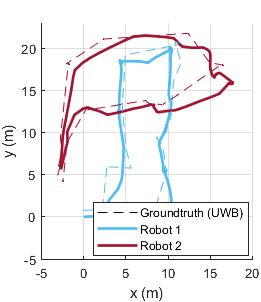}
            \caption{Hardware experiment results. Left: optimized trajectory from Disco-SLAM without using \textit{Wi-Closure}. Right: optimized trajectory from Disco-SLAM using \textit{Wi-Closure.}}
            \label{fig:hardware_traj}
        \end{figure}

\begin{table}[H]
\centering
\begin{tabular}{ |p{4cm}|p{1.5cm}|p{1.5cm}|}
 \hline
   & \textbf{Baseline} &  \textbf{\textit{Wi-Closure}}\\
 \hline
 ATE (m) & 17.6 & 1.9 \\
 \hline
%  Correctly included true LC & 5544 & 3631 \\
%  \hline
 Correctly rejected false LC (\%) & N/A & 78.7 \\
 \hline
%  Incorrectly included false LC & 1093557 & 3418 \\
%  \hline
 Missed true LC (\%) & 0 & 15 \\
 \hline
 Total Computation Time (s) & 155 & 36 \\
 \hline
 Total Wi-Closure time (s) & N/A & 0.5 seconds \\
 \hline
\end{tabular}
\caption{Loop Closure (LC) performance comparison between \textit{Wi-Closure} and DiSCO-SLAM in hardware experiments.}
\label{tab:hardware_results}
\end{table}
\subsection{Hardware experiments}
We evaluate our approach on a dataset collected in an unfinished shell space as shown in Fig.~\ref{fig:shellspace} with repetitive features. We deploy two customized Locobot PX100, which are installed with a Velodyne VLP-16 LiDAR, a MicroStrain 3DM-GX5-AHRS IMU, DWM1001 UWB, 5dBi Antenna and Intel NUC 10. We process the AOA measurements using the WiFi sensing Toolbox from our earlier work \cite{Jadhav2021}. To accompany the scale of the test field, we limit the range of the LiDAR to 10 meters. For the purpose of computing the ground truth error, we set up 5 UWB nodes in the space to localize the robot in real-time.\\
Two robots are set up at different locations without knowing each other's frames. They traverse the space collecting LiDAR scans and IMU data. Every 10 meters one robot collects AOA and ranging measurements to the other robot. Trajectories have two rendezvous points to provide loop closure opportunities. Again, we compare computation time and ATE with and without \textit{Wi-Closure}, and we assess if loop closures are filtered correctly. %Without losing generalizability, we don't specifically tune the parameters of DiSCO SLAM to better fit our dataset but rather 

We directly apply original DiSCO-SLAM parameters from \cite{Huang2022}, and show that the original method fails in our environment while adding \textit{Wi-Closure} recovers performance. \\
As shown in Table~\ref{tab:hardware_results}, our approach successfully reduces computation time of the whole SLAM pipeline by 4.3 times and reduces the trajectory error by 89.2\%. Fig.~\ref{fig:hardware_traj} shows the optimized trajectories. Because of the repetitiveness of the pillars, the original algorithm fails in the challenging environment. Similar to the simulation results, applying our approach substantially reduces the search space from 1,848 loop closures to only 119 of which 115 are true inter-robot loop closures. Consequently, our method increases speed and prevents failure of the algorithm. \\
Also, our approach successfully handles the multipath phenomenon in our hardware experiment. Each of four AOA measurements contains five multipath. \textit{Wi-Closure} is able to distinguish all three direct path from the 17 multipath, leading to consistent optimization results as shown in Fig.~\ref{fig:hardware_traj}.

\section{Conclusion}
In this paper we propose an efficient and robust loop closure finding method \textit{Wi-Closure}, utilizing light-weight information from the wireless signal between robots. We properly handle the multipath phenomenon, and are able to exclude the majority of false loop closures. This drastically reduces processing time of the muli-robot SLAM pipeline and increases the robustness of the results.

% \begin{acks}
% We gratefully acknowledge the funding support through the NSF CAREER reward CNS-2114733, Young Investigator Program(YIP) N00014-21-1-2714 of the Office of Naval Research (ONR), and U.S. Office of Naval Research Global (ONRG) NICOP-grant N62909-19-1-2027.
% \end{acks}

\section{Acknowledgement}
We gratefully acknowledge partial funding support through the Amazon Research Awards (ARA), the Office of Naval Research (ONR) Young Investigator Program(YIP) N00014-21-1-2714, and U.S. Office of Naval Research Global (ONRG) NICOP grant N62909-19-1-2027.

%------- CITATIONS ----------
% \nocite{}
\bibliography{references.bib}

\begin{thebibliography}{10}
\providecommand{\url}[1]{#1}
\csname url@samestyle\endcsname
\providecommand{\newblock}{\relax}
\providecommand{\bibinfo}[2]{#2}
\providecommand{\BIBentrySTDinterwordspacing}{\spaceskip=0pt\relax}
\providecommand{\BIBentryALTinterwordstretchfactor}{4}
\providecommand{\BIBentryALTinterwordspacing}{\spaceskip=\fontdimen2\font plus
\BIBentryALTinterwordstretchfactor\fontdimen3\font minus
  \fontdimen4\font\relax}
\providecommand{\BIBforeignlanguage}[2]{{%
\expandafter\ifx\csname l@#1\endcsname\relax
\typeout{** WARNING: IEEEtran.bst: No hyphenation pattern has been}%
\typeout{** loaded for the language `#1'. Using the pattern for}%
\typeout{** the default language instead.}%
\else
\language=\csname l@#1\endcsname
\fi
#2}}
\providecommand{\BIBdecl}{\relax}
\BIBdecl

\bibitem{angeli2008fast}
A.~Angeli, D.~Filliat, S.~Doncieux, and J.-A. Meyer, ``Fast and incremental
  method for loop-closure detection using bags of visual words,'' \emph{IEEE
  transactions on robotics}, vol.~24, no.~5, pp. 1027--1037, 2008.

\bibitem{hess2016real}
W.~Hess, D.~Kohler, H.~Rapp, and D.~Andor, ``Real-time loop closure in 2d lidar
  slam,'' in \emph{2016 IEEE international conference on robotics and
  automation (ICRA)}.\hskip 1em plus 0.5em minus 0.4em\relax IEEE, 2016, pp.
  1271--1278.

\bibitem{bowman2017probabilistic}
S.~L. Bowman, N.~Atanasov, K.~Daniilidis, and G.~J. Pappas, ``Probabilistic
  data association for semantic slam,'' in \emph{2017 IEEE international
  conference on robotics and automation (ICRA)}.\hskip 1em plus 0.5em minus
  0.4em\relax IEEE, 2017, pp. 1722--1729.

\bibitem{dube2017online}
R.~Dub{\'e}, A.~Gawel, H.~Sommer, J.~Nieto, R.~Siegwart, and C.~Cadena, ``An
  online multi-robot slam system for 3d lidars,'' in \emph{2017 IEEE/RSJ
  International Conference on Intelligent Robots and Systems (IROS)}.\hskip 1em
  plus 0.5em minus 0.4em\relax IEEE, 2017, pp. 1004--1011.

\bibitem{Mangelson2018}
J.~G. Mangelson, D.~Dominic, R.~M. Eustice, and R.~Vasudevan, ``Pairwise
  consistent measurement set maximization for robust multi-robot map merging,''
  in \emph{2018 IEEE International Conference on Robotics and Automation
  (ICRA)}, 2018, pp. 2916--2923.

\bibitem{Ikram2022}
M.~H. Ikram, S.~Khaliq, M.~L. Anjum, and W.~Hussain, ``Perceptual aliasing++:
  Adversarial attack for visual slam front-end and back-end,'' \emph{IEEE
  Robotics and Automation Letters}, vol.~7, no.~2, pp. 4670--4677, 2022.

\bibitem{Kaess2019}
M.~Hsiao and M.~Kaess, ``Mh-isam2: Multi-hypothesis isam using bayes tree and
  hypo-tree,'' in \emph{2019 International Conference on Robotics and
  Automation (ICRA)}, 2019, pp. 1274--1280.

\bibitem{Shienman2022}
M.~Shienman and V.~Indelman, ``D2a-bsp: Distilled data association belief space
  planning with performance guarantees under budget constraints,'' 05 2022, pp.
  11\,058--11\,065.

\bibitem{Kumar2014AccurateIL}
S.~Kumar, S.~Gil, D.~Katabi, and D.~Rus, ``Accurate indoor localization with
  zero start-up cost,'' in \emph{MobiCom '14}, 2014.

\bibitem{Jadhav2021}
\BIBentryALTinterwordspacing
N.~Jadhav, W.~Wang, D.~Zhang, S.~Kumar, and S.~Gil, ``Toolbox release: A
  wifi-based relative bearing sensor for robotics,'' 2021. [Online]. Available:
  \url{https://arxiv.org/abs/2109.12205}
\BIBentrySTDinterwordspacing

\bibitem{Olson2011tags}
E.~Olson, ``{AprilTag}: A robust and flexible visual fiducial system,'' in
  \emph{Proceedings of the {IEEE} International Conference on Robotics and
  Automation ({ICRA})}.\hskip 1em plus 0.5em minus 0.4em\relax IEEE, May 2011,
  pp. 3400--3407.

\bibitem{GalvezLopez2012}
D.~Galvez-López and J.~D. Tardos, ``Bags of binary words for fast place
  recognition in image sequences,'' \emph{IEEE Transactions on Robotics},
  vol.~28, no.~5, pp. 1188--1197, 2012.

\bibitem{Kin2006}
\BIBentryALTinterwordspacing
K.~L. Ho and P.~Newman, ``Loop closure detection in slam by combining visual
  and spatial appearance,'' \emph{Robotics and Autonomous Systems}, vol.~54,
  no.~9, pp. 740--749, 2006, selected papers from the 2nd European Conference
  on Mobile Robots (ECMR ’05). [Online]. Available:
  \url{https://www.sciencedirect.com/science/article/pii/S0921889006000844}
\BIBentrySTDinterwordspacing

\bibitem{Giamou2018}
M.~Giamou, K.~Khosoussi, and J.~P. How, ``Talk resource-efficiently to me:
  Optimal communication planning for distributed loop closure detection,'' in
  \emph{2018 IEEE International Conference on Robotics and Automation (ICRA)},
  2018, pp. 3841--3848.

\bibitem{song2019uwb}
Y.~Song, M.~Guan, W.~P. Tay, C.~L. Law, and C.~Wen, ``Uwb/lidar fusion for
  cooperative range-only slam,'' in \emph{2019 international conference on
  robotics and automation (ICRA)}.\hskip 1em plus 0.5em minus 0.4em\relax IEEE,
  2019, pp. 6568--6574.

\bibitem{uwb_multi1}
A.~Fishberg and J.~P. How, ``{Multi-Agent} relative pose estimation with {UWB}
  and constrained communications,'' Mar. 2022.

\bibitem{uwb_multi2}
E.~R. Boroson, R.~Hewitt, N.~Ayanian, and J.-P. de~la Croix, ``{Inter-Robot}
  range measurements in pose graph optimization,'' in \emph{2020 {IEEE/RSJ}
  International Conference on Intelligent Robots and Systems ({IROS})}, Oct.
  2020, pp. 4806--4813.

\bibitem{arraywifi}
J.~Xiong and K.~Jamieson, ``Arraytrack: A fine-grained indoor location
  system,'' in \emph{Proceedings of the 10th USENIX Conference on Networked
  Systems Design and Implementation}, ser. nsdi'13.\hskip 1em plus 0.5em minus
  0.4em\relax USA: USENIX Association, 2013, p. 71–84.

\bibitem{rangeonly}
E.~R. Boroson, R.~Hewitt, N.~Ayanian, and J.-P. de~la Croix, ``Inter-robot
  range measurements in pose graph optimization,'' in \emph{2020 IEEE/RSJ
  International Conference on Intelligent Robots and Systems (IROS)}, 2020, pp.
  4806--4813.

\bibitem{ORB}
E.~Rublee, V.~Rabaud, K.~Konolige, and G.~Bradski, ``Orb: An efficient
  alternative to sift or surf,'' in \emph{2011 International Conference on
  Computer Vision}, 2011, pp. 2564--2571.

\bibitem{Tian2021b}
\BIBentryALTinterwordspacing
Y.~Tian, K.~Khosoussi, and J.~P. How, ``A resource-aware approach to
  collaborative loop-closure detection with provable performance guarantees,''
  \emph{The International Journal of Robotics Research}, vol.~40, no. 10-11,
  pp. 1212--1233, 2021. [Online]. Available:
  \url{https://doi.org/10.1177/0278364920948594}
\BIBentrySTDinterwordspacing

\bibitem{Hsiao2019}
M.~Hsiao and M.~Kaess, ``Mh-isam2: Multi-hypothesis isam using bayes tree and
  hypo-tree,'' in \emph{2019 International Conference on Robotics and
  Automation (ICRA)}, 2019, pp. 1274--1280.

\bibitem{Pathak2016}
S.~Pathak, A.~Thomas, A.~Feniger, and V.~Indelman, ``Da-bsp: Towards data
  association aware belief space planning for robust active perception,'' in
  \emph{ECAI}, 2016.

\bibitem{Cadena2016}
C.~Cadena, L.~Carlone, H.~Carrillo, Y.~Latif, D.~Scaramuzza, J.~Neira, I.~Reid,
  and J.~J. Leonard, ``Past, present, and future of simultaneous localization
  and mapping: Toward the robust-perception age,'' \emph{IEEE Transactions on
  Robotics}, vol.~32, no.~6, pp. 1309--1332, 2016.

\bibitem{Naseri2019}
H.~Naseri and V.~Koivunen, ``A bayesian algorithm for distributed network
  localization using distance and direction data,'' \emph{IEEE Transactions on
  Signal and Information Processing over Networks}, vol.~5, no.~2, pp.
  290--304, 2019.

\bibitem{Huang2022}
Y.~Huang, T.~Shan, F.~Chen, and B.~Englot, ``Disco-slam: Distributed scan
  context-enabled multi-robot lidar slam with two-stage global-local graph
  optimization,'' \emph{IEEE Robotics and Automation Letters}, vol.~7, no.~2,
  pp. 1150--1157, 2022.

\end{thebibliography}
\bibliographystyle{IEEEtran}

\end{document}